\pdfoutput=1 
\documentclass{article}
\usepackage{algorithm}
\usepackage{algorithmic}
\usepackage{color}
\usepackage{sidecap}
\usepackage{subcaption}
\usepackage{wrapfig}
\usepackage{graphicx}
\usepackage{amsmath}
\usepackage{dsfont}
\usepackage{lipsum}
\usepackage{footnote}
\newcommand\blfootnote[1]{%
  \begingroup
  \renewcommand\thefootnote{}\footnote{#1}%
  \addtocounter{footnote}{-1}%
  \endgroup
}

\usepackage[final]{corl_2018}
\title{GPU-Accelerated Robotic Simulation for Distributed Reinforcement Learning}

\author{
  Jacky Liang$^{\dag \ddag 1}$ \\
  Robotics Institute \\
  Carnegie Mellon University \\
  \And
  Viktor Makoviychuk$^{\dag 2}$ \\
  NVIDIA \\
  \And
  Ankur Handa$^{\dag 2}$ \\
  NVIDIA \\
  \And
  Nuttapong Chentanez$^2$ \\
  NVIDIA \\
  \And
  Miles Macklin$^2$ \\
  NVIDIA \\ University of Copenhagen \\
  \And
  Dieter Fox$^2$ \\
  NVIDIA \\
}
\makesavenoteenv{tabular}
\makesavenoteenv{table}
\begin{document}
\maketitle
{\centering
$^1$\texttt{jackyliang@cmu.edu}\\
$^2$\texttt{\{vmakoviychuk, ahanda, nchentanez, mmacklin, dieterf\}@nvidia.com}\par
}

\blfootnote{$^{\dag}$ Shared first author}
\blfootnote{$^{\ddag}$ Work done during internship at NVIDIA}

\begin{abstract}
Most Deep Reinforcement Learning (Deep RL) algorithms require a prohibitively large number of training samples for learning complex tasks. 
Many recent works on speeding up Deep RL have focused on distributed training and simulation.
While distributed training is often done on the GPU, simulation is not. 
In this work, we propose using GPU-accelerated RL simulations as an alternative to CPU ones.
Using NVIDIA Flex, a GPU-based physics engine, we show promising speed-ups of learning various continuous-control, locomotion tasks.
With one GPU and CPU core, we are able to train the Humanoid running task in less than 20 minutes, using $10-1000\times$ fewer CPU cores than previous works.
We also demonstrate the scalability of our simulator to multi-GPU settings to train more challenging locomotion tasks.
\end{abstract}

\keywords{Deep Reinforcement Learning, GPU Acceleration, Simulation} 
\section{Introduction}
\label{sec:introduction}

Model-free Deep RL has seen impressive achievements \cite{mnih2016asynchronous, Silver_AlphaGo, schulman2017proximal} in recent years, but many methods and tasks require enormous amount of compute due to the large sample complexity of exploration in high-dimensional state and action spaces. 
One approach to overcome exploration is by using human demonstrations \cite{HesterVPLSPSDOA17,Rajeswaran:etal:RSS2018}, but collecting human demonstrations remain challenging for many tasks, and it is difficult to scale.
Another approach is to vastly scale up RL simulation and training to distributed settings, so large amounts of data can be obtained in a relatively short amount of time.
Many recent works of this approach have seen scaling benefits by performing policy training on the GPU while scaling up environment simulation on many CPUs.
In this work, we propose using a GPU-accelerated RL simulator to bring the benefits of GPU's parallelism to RL simulation as well.

Using Flex, a GPU-based physics engine developed with CUDA, we implement an OpenAI Gym-like interface to perform RL experiments for continuous control locomotion tasks.
We benchmark our simulator on ant and humanoid running tasks as well as their more challenging variations, inspired by ones proposed in OpenAI Roboschool and the Deepmind Parkour environments.
They include learning to run toward changing target locations, recovering from falls, and running on complex, uneven terrains.
Our choice of tasks is driven by their popularity and the challenges they offer to various Deep RL algorithms.
Although our training results are not directly comparable to those obtained in physics simulators used in prior work (\textit{e.g.} MujoCo, Bullet) due to differences in physics simulation, we have endeavoured to do head-to-head comparisons wherever possible.
Using our GPU-accelerated RL framework to simulate and train hundreds to thousands of agents at once on a single GPU, we were able to achieve faster training results than previous works which used large CPU clusters.
In addition, the scale and speed-ups achieved through our simulator, especially in the more challenging tasks, make GPU-accelerated RL simulation a viable alternative to CPU ones.

\begin{figure*}[t]
\vspace{2mm}
\centerline{
\hfill { \includegraphics[width=0.33\linewidth]{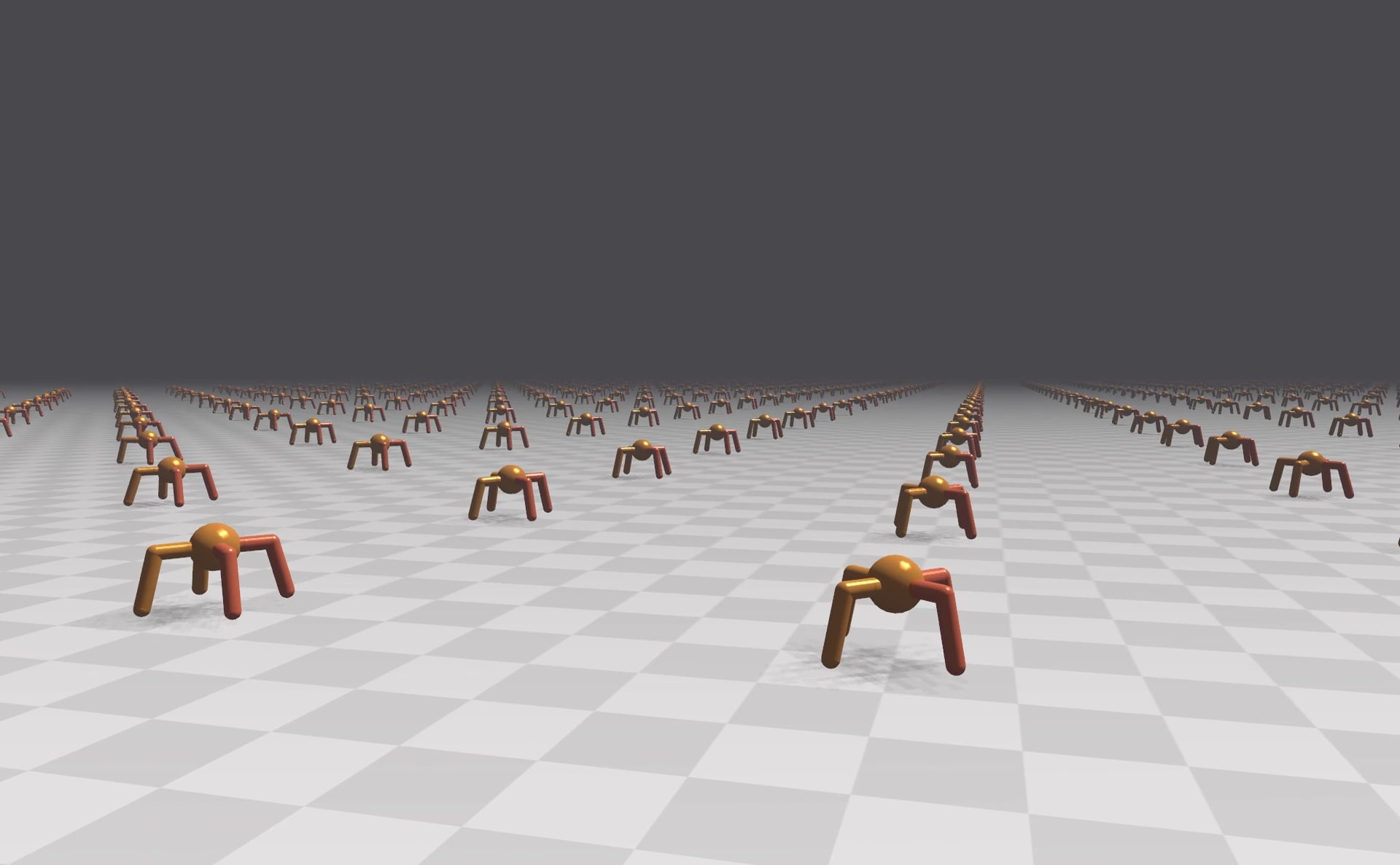} }
\hfill { \includegraphics[width=0.33\linewidth]{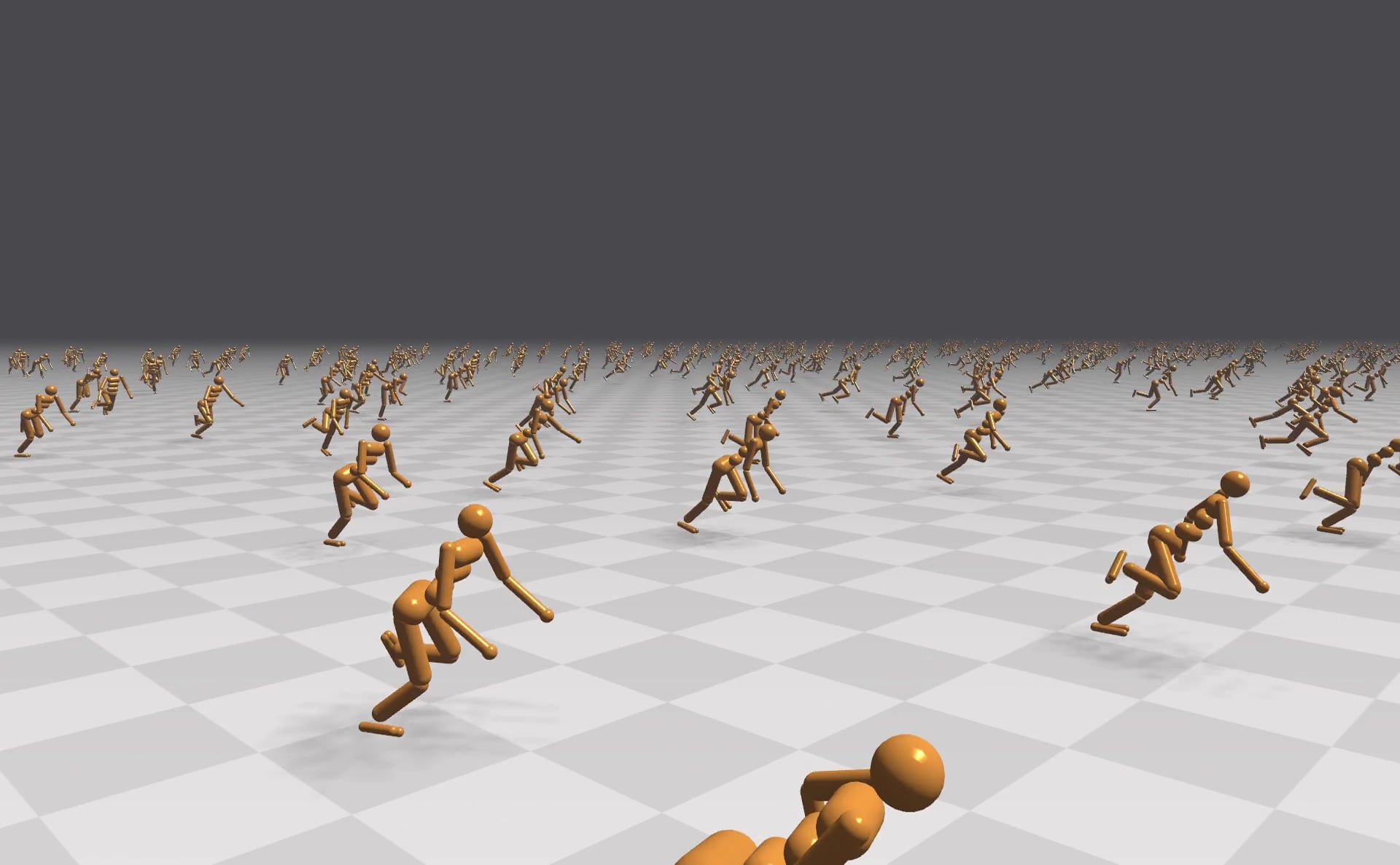} }
\hfill { \includegraphics[width=0.33\linewidth]{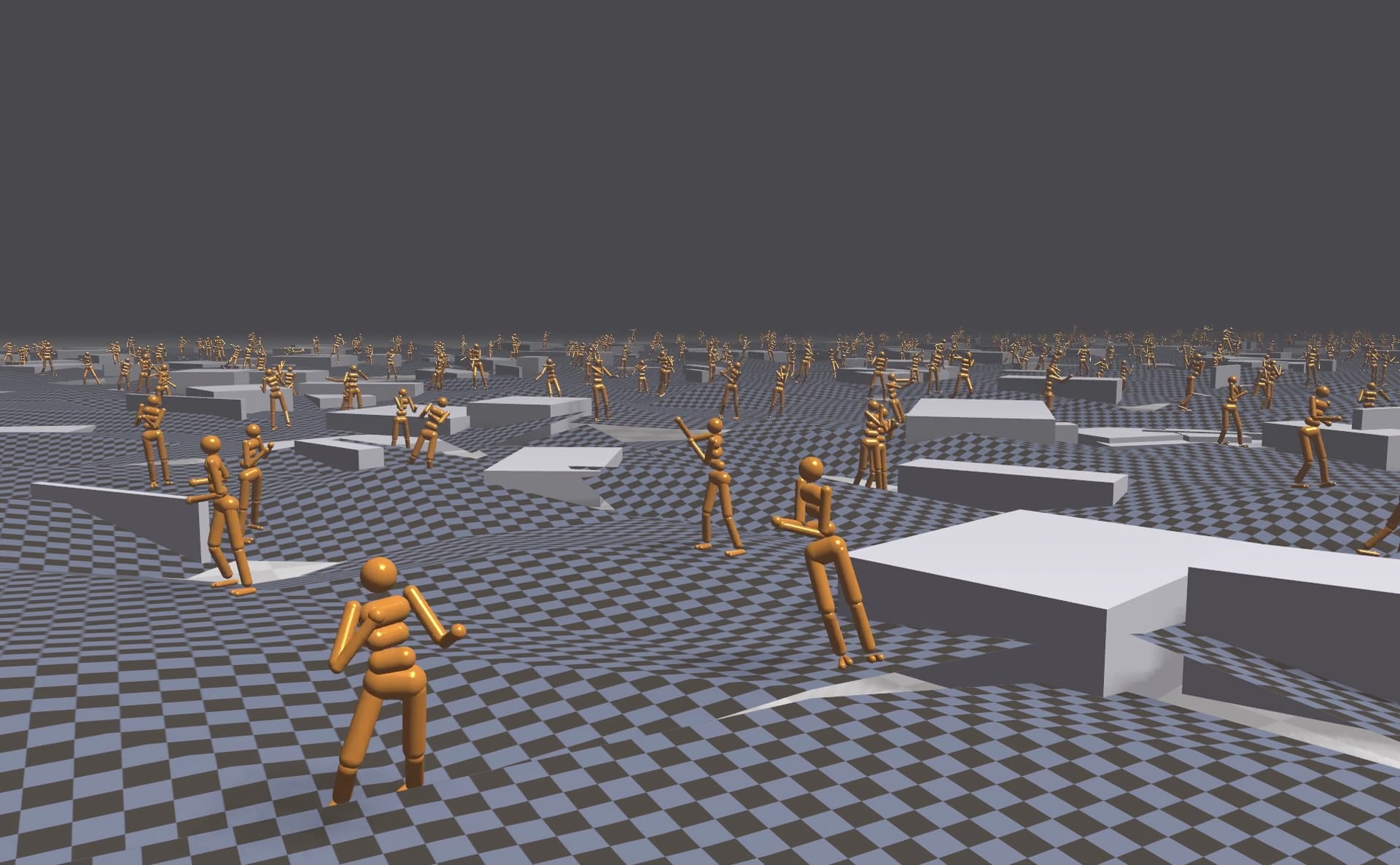} }
\hfill
}
\caption{\footnotesize{GPU-Accelerated RL Simulation. We use an in-house GPU-accelerated physics simulator, to concurrently simulate hundreds to thousands of robots for Deep RL of continuous control locomotion tasks. Here we show the Ant, Humanoid, and Humanoid Flagrun Harder on Complex Terrain tasks benchmarked in our work. Using a single machine (1 GPU and CPU core), we are able to train humanoids to run in less than 20 minutes.}}
\label{fig:screenshots}
\end{figure*}


We summarize our key contributions below:
\begin{enumerate}
    \item A GPU-accelerated RL simulator built with an in-house GPU-based physics engine.  We plan to release our simulator in the near future to facilitate fast RL training to the community. 
    \item Experiments on massively distributed simulations of hundreds to thousands of locomotion environments on single and multiple GPU settings.
    \item Improvements in training speed for various challenging locomotion tasks, learning the humanoid running task in less than 20 minutes on a single machine. See our trained policies at \url{https://sites.google.com/view/accelerated-gpu-simulation/home}.
\end{enumerate}
We note that in this paper our focus is on the application of our GPU-based physics engine to RL simulation, and not on comparisons and benchmarks of the physics engine itself. 

\section{Related Works}
\label{sec:related-works}

\subsection{Distributed Deep RL}

Many prior works have explored parallelizing environment simulation and policy training on CPUs. 
\citet{nair2015massively} proposed the first massively parallelized method for training RL agents. 
Their method, Gorila DQN, has separate learners, actors, and parameter servers. Simulating and training with hundreds of CPU cores it was able to achieve superhuman performance in most Atari games in a few days.
Following Gorila DQN, \citet{mnih2016asynchronous} proposed the Asynchronous-Actor-Critic-Agents (A3C) algorithm, and with 16 CPU cores they were able to compete with Gorila DQN in Atari games with about the same training time.

\citet{babaeizadeh2016reinforcement} extended A3C by proposing a CPU/GPU hybrid variation of A3C, GA3C, which moves the policy to the GPU. 
This enabled GPU-accelerated policy inference and training. 
Their algorithm dynamically adjusts the number of parallel agents, trainers, and parameter servers to maximize training speed.
On a single machine with 16 CPU cores and 1 Tesla K40 GPU GA3C achieved more than $4\times$ speed-up over a CPU-only implementation of A3C.
\citet{adamski2018distributed} furthered scaled up A3C training by using larger batchsizes and a well-tuned Adam Optimizer. 
Their method allowed them to learn many Atari games with hundreds of CPU cores, with no GPU, in less than an hour (\textit{e.g.} Breakout with $768$ CPU cores in $21$ minutes). 

\citet{salimans2017evolution} explored using evolutionary strategies (ES) for RL.
The authors scaled up ES to use as much as $720$ CPU cores to learn Atari games in about one hour.
They also performed learning experiments with MuJoCo locomotion tasks, and with $1440$ CPU cores they were able to train a humanoid to walk in $10$ minutes.
\citet{such2017deep} applied Genetic Algorithms (GA), a gradient-free optimization method, to RL. 
GA is also amenable to parallelization, and it was able to learn Atari games also in one hour with $720$ CPU cores.
In humanoid locomotion tasks however, the algorithm trained much slower than ES, and was not able to achieve comparable performance in the same time frame. 

Recent advances in parallel computation tools such as Horovod~\citep{sergeev2018horovod} and Ray have enabled researchers to easily scale up machine learning to distributed settings.
Ray RLLib~\citep{liang2017ray} is a distributed RL library using the Ray framework.
In their benchmarks, the authors were able to scale ES with Ray RLLib to more than $8000$ CPU cores, learning the humanoid walking task in just $3.7$ minutes. 
\citet{mania2018simple} also used Ray but for their proposed algorithm, Augmented Random Search (ARS).
ARS learned the humanoid walking task in $21$ minutes with $48$ CPU cores, while using $15\times$ less CPU time than ES. 

Other previous works aimed to improve the efficiency of off-policy learning from the large amount of data generated by many parallel actors.
\citet{espeholt2018impala} proposed IMPALA, which applies large-scale, distributed learning systems to solving multi-task RL problems.
IMPALA is inspired by A3C, but the actors don't compute and send policy gradients to the learners - they send the sampled trajectories instead.
The authors scaled IMPALA to use $500$ CPU actors and $8$ GPU learners, and it learned the benchmarked tasks (DMLab-30, a suite of multi-task video game-like environments) in less than a day. 
\citet{horgan2018distributed} introduced Distributed Prioritized Experience Replay, an off-policy RL algorithm that uses a novel technique to sample more important trajectories in its replay buffer for learning.
This work uses many CPU cores for simulating the RL environment, and 1 GPU for training.
With $360$ actors, the method learned most Atari games in a few hours, and with $32$ actors it trained a humanoid to walk in $1$ hour, run in $4$ hours. 
\citet{stooke2018accelerated} explored optimizing existing RL algorithms for fast performance on a multi-GPU system. 
The authors used an entire NVIDIA DGX-1, which contains $8$ NVIDIA Tesla V100 GPUs. 
Running $256$ parallel simulations on $40$ CPU cores and performing training on all $8$ V100s, the authors report being able to train many Atari games in less than $10$ minutes.
Recently, OpenAI massively scaled up Proximal Policy Optimization~\citep{schulman2017proximal} to use more than $6000$ CPU cores to simulate in-hand manipulation tasks~\citep{openai2018dexterous} and more than $100,000$ for playing Dota~\footnote{\url{https://openai.com/five/}}. 

\subsection{Locomotion and Physics Simulation}

We focus our attention to continuous control locomotion tasks first proposed in MuJoCo~\citep{6386109}, a popular CPU-based physics simulator. 
DART and Bullet are other notable alternatives, but MuJoCo remains by far the most popular physics simulator in the Deep RL community ~\citep{salimans2017evolution, such2017deep, liang2017ray, mania2018simple, horgan2018distributed, duan2016benchmarking, schulman2017proximal, heess2017emergence} for its efficient simulation and relatively stable contact models. 
\citet{duan2016benchmarking} first benchmarked different RL algorithms on various continuous control tasks such as cartpole swing-up and humanoid walking forward. Later, \citet{schulman2017proximal} introduced more complex locomotion tasks, such as humanoid flagrun where the agent must learn to change and run toward different directions. 
\citet{heess2017emergence} take this one step further and train humanoid agents to walk and run on uneven and dynamic terrains. 
Taking inspiration from these works, we use the humanoid running task and its more challenging variations for benchmarking.
Owing to the humanoid's high degree of freedom control space, its tasks require most Deep RL algorithms to use a significant number of samples to learn, which provide opportunities for improving learning speed via reduction in simulation time.

Many previous works on distributed RL have focused on discrete control problems such as Atari games, which do not require physics simulation. 
Moreover, the works in continuous control tasks have only used CPU-based simulations. 
While GPU-accelerated physics simulations have been applied in scientific computing~\citep{freniere2016feasibility, spiechowicz2015gpu} and healthcare~\citep{blumers2017gpu, wu2015software}, they have yet to be applied in robotics.
To achieve state-of-the-art performance, previous works often had to scale environment simulation to hundreds, if not thousands of CPU cores. 
In our work, we explore using GPU-accelerated simulation as an alternative to CPU-based ones. 
Using a single GPU, we can simulate hundreds to thousands of robots and achieve state-of-the-art results in locomotion tasks on a single machine, learning the humanoid running task in less than 20 minutes. 
\section{GPU-Accelerated Robotics Simulation}
\label{sec:gpu-accelerated-robotics-simulation}

\subsection{GPU-based Physics Engine}
Our in-house GPU-based physics engine uses a non-smooth Newton method for its rigid body solver and a maximal coordinate formulation.
Like with environments in MuJoCo and Bullet, we use torque control as the actuation model.
Potential collisions and contacts among bodies are detected speculatively and are modeled through unilateral constraint functions with a smooth isotropic Coulomb friction model.
We use sliding friction coefficient of 1.0, the same as in MuJoCo~\citep{todorov2010implicit}.
Restitution coefficient is $0.0$ and gravity is $9.8\frac{m}{s^2}$ downward.
For time-stepping we use an implicit time-discretization also like~\citep{todorov2010implicit}, and the time step used is $\frac{1}{120}$s.
Each Newton iteration is solved by a sparse iterative Krylov linear solver with the minimum number of linear iterations such that simulation is stable for our experiments.
We found Krylov methods allowed sufficient stiffness to achieve realistic humanoid gaits, while relaxation methods like Projected Gauss-Seidel were less effective, especially when paired with a maximal coordinate representation. 

We develop a GPU-accelerated robotics simulation framework for RL that can simulate many robot agents in parallel for a variety of tasks. 
To simulate multiple robots performing the same task in parallel, we load all robots and task-related objects into the \textit{same} simulation.
This is unlike previous works that parallelize simulation by using multiple CPU processes or threads, each simulating an individual environment.
The parallelism and the speed-up in our simulation is achieved by performing the physics computations on the GPU.
We note that in our simulations, agents are able to interact with each other, which is not possible for running multiple simulation processes of 1 agent each. 


\subsection{GPU Simulation Speed}
\begin{figure}
    \centering
    \includegraphics[width=0.45\linewidth]{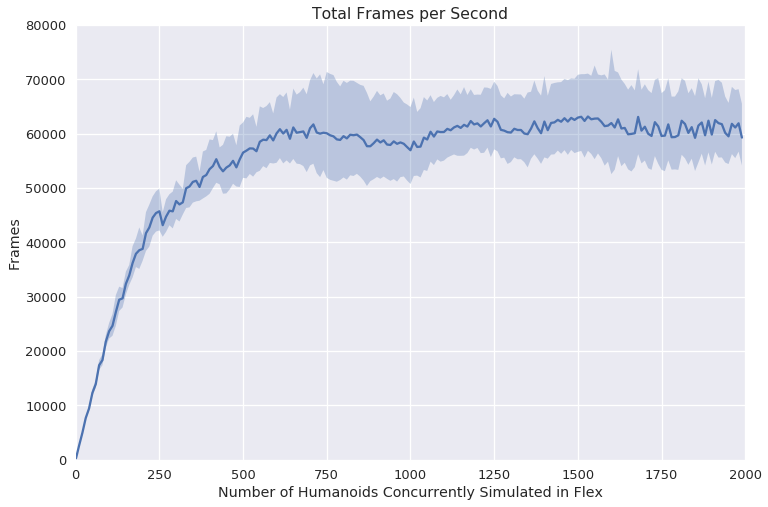}
    \includegraphics[width=0.45\linewidth]{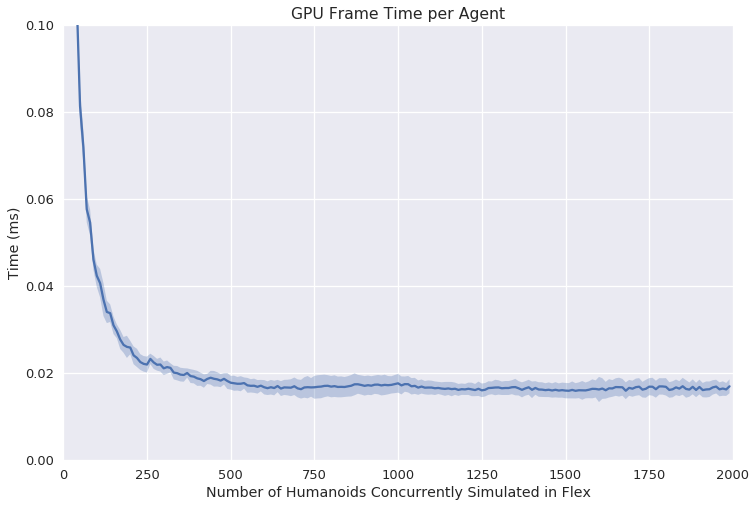}
    \caption{\footnotesize GPU Simulation Speed. We measure the speed of GPU simulation for the humanoid task as we increase the number of concurrent humanoids simulated. The total simulations per second peaks at around $60$KHz for $750$ humanoids, and the best mean GPU simulation frame time per agent is less than $0.02 ms$. The simulation time grows much slower than the number of humanoids because of the constant CUDA kernels launch overhead, which dominates in total step time when only a few humanoids are available.}
    \label{fig:sim_times}
\vspace{-4mm}
\end{figure}

To illustrate the typical performance of our simulator on a single-machine setting, we measured the GPU simulation frame time for the humanoid task as we increase the number of humanoids concurrently simulated in the environment.
The results are obtained on an NVIDIA Tesla V100 GPU.
The GPU simulation frame time does not include the time needed for calculating rewards, applying actions, and transfer RL-related data back and forth from the Python client, as that speed varies based on implementation of the RL framework.
In Figure~\ref{fig:sim_times} we report two plots, the total simulation frames generated per second, calculated by multiplying the number of agents by the frames per second of the entire simulation, and the GPU frame time per agent.
We note that both values converge with around $750$ simulated humanoids, where it can generate $60$K frames per second, and the mean frame time per agent is below $0.02$ms. 

We observe in our learning experiments that although the total simulation frames generated per second peaks around $750$ agents, this is not the optimal number of agents for minimizing learning speed.
The number of agents used affects how the learning algorithm and policy explores the state and action spaces, and the threshold for optimizing learning speed depends on the specific task. 

\section{Experiments}
\label{sec:experiments}

To evaluate the performance of our GPU-accelerated physics simulator for RL, we perform a set of learning experiments on various locomotion tasks. 
We first measure the learning performance on a single GPU as we vary the number of parallel agents simulated on that GPU.
Then we measure how well learning performance scales with multiple GPUs and nodes as we fix the number of agents simulated per GPU.

\subsection{Tasks}
We perform learning experiments on the following 4 tasks, 3 of which are shown in Figure~\ref{fig:screenshots}. 

\textbf{Ant.}
We use the ant model commonly found in MuJoCo locomotion environments.
It has $4$ legs and $8$ controllable joints that form the action space. 
The goal of the Ant task is to have the agent move forward as fast as possible.
Ant is relatively easy to learn, because the initial state of the task is stable.
This makes it a useful task for sanity checks and debugging.

\textbf{Humanoid.}
Like~\citep{heess2017emergence, tassa2018deepmind}, we use the humanoid model with 28 degrees of freedom and 21 actuated joints.
This humanoid is more complex than the 24-DoF humanoid model with 17 actuated joints used in~\citep{schulman2017proximal, salimans2017evolution, liang2017ray, mania2018simple}.
The 28-DoF humanoid has 4 additional joints to control the ankle angles of the robot, whereas the 24-DoF one has ball-shaped feet that cannot be rotated.
We choose the more complex humanoid for benchmarking, because the additional ankle joints allow the humanoid to learn more realistic running and turning motions.
The goal of the Humanoid task is to have the agent move forward as fast as possible.
This task is often used in the literature for benchmarking locomotion learning performance. 

The observations for Ant and Humanoid include the agent's height, velocity, and joint angles, among others.
See Appendix~\ref{appendix:obs_comp} for a detailed comparison of observations used in our and previous works.

\textbf{Humanoid Flagrun Harder (HFH).}
In the HFH task, a humanoid must learn to not only run forward but also to turn and run toward different target locations.
The agent must also learn how to recover from falling by standing up.
This is a much more challenging task than vanilla Humanoid, and it takes more time to train.
The action space of this task is the same as that in the Humanoid task.
We observe that training a humanoid for HFH leads to more robust and symmetric walking gaits \textit{e.g.} humanoids can maintain their stand-up and running skills even with as much as 50\% higher or lower gravity. 

\textbf{Humanoid Flagrun Harder on Complex Terrain.}
In this task, the agent must learn to run and change directions on uneven terrain with static rectangular obstacles.
The dimensions, location, and orientation of the obstacles are randomly sampled from a uniform distribution. 
The action space of this task is the same as that in the Humanoid task.
To help the humanoid navigate complex terrain and overcome obstacles, we augment the observation space with a 2D, $15 \times 11$ rectangular height map that follows the humanoid's center of mass.
Similar to~\citep{heess2017emergence}, our height map is denser near the humanoid.

For Ant and Humanoid, an episode terminates if the agents fall below a threshold.
For the two HFH tasks, we allow the agents to fall below a threshold for a certain time period (160 frames) before terminating the agent.
This enables the agents to learn to stand up and recover from a fall.
The Flagrun targets for the HFH tasks change every 200 frames to a random location within $100m$ of the agent, or earlier if the humanoid reaches within $1m$ of the target. 
For all tasks, the maximum episode length for both training and evaluation is $1000$ frames.

\textbf{Rewards.}
Similar to previous works, the reward function for all tasks reward the current speed toward the desired targets and penalize excessive torque applied to the joints.
Our reward functions however, are not immediately comparable to previous works, due to a smaller alive bonus and the addition of other terms that we empirically found to lead to more natural, symmetric running gaits. 
We note that the locomotion rewards used in MuJoCo and Bullet are also different, arising from the vagaries of implementation details, such as the solver and the number of iterations used for optimisation.
See Appendix~\ref{appendix:rewards}, \ref{appendix:reward_comp} for the exact rewards used and a comparison of our rewards with those used in previous work.

A common reward threshold for solving the humanoid running forward task is $6000$~\citep{salimans2017evolution, liang2017ray, mania2018simple}, but this threshold is for the 24-DoF humanoid, and to our knowledge there is no widely used reward threshold for the 28-DoF humanoid and for the HFH tasks.
For the Humanoid task, we chose a reward threshold of 3000 for walking and $5000$ for running.
$3000$ roughly corresponds to a forward moving speed of $2$~\textit{m/s}, and $5000$ for $4$~\textit{m/s}, which is about the same speed as Roboschool's example 24-DoF humanoid running policy.
For Ant, we use $3000$ as the reward threshold for running, and $7000$ for final reward.
For the HFH task, we use $2500$ as an intermediary reward threshold, around which the agents first learn to stand up from the ground, and $4000$ as the final reward. 

\textbf{Initial State and Perturbations.}
Unlike parallel simulations on CPUs that simulate multiple agents in their own environments, usually one environment per CPU core, we simulate all agents in one environment on the GPU.
The initial positions, velocities, joint angles, and joint angular velocities of agents are perturbed slightly during initialization.
We also exert random forces onto the agents for all 4 tasks every $200$ to $300$ frames, for a few Newtons each time. 
These external perturbations help the agent to learn more robust policies.
We also enabled inter-agent collisions for the HFH tasks.
The initial spacing of the humanoids affect the collision frequency, and the occasional collisions help the agents to explore states where they must learn to balance and recover from falls, leading to more robust policies.

\subsection{Learning Algorithm}
We use a variation of Proximal Policy Optimization (PPO)~\citep{schulman2017proximal} for all our experiments to benchmarks locomotion tasks. 
We adapted the open source OpenAI Baseline~\footnote{\url{https://github.com/openai/baselines}} implementation of PPO to work with our simulation environment, where a single environment step simulates multiple agents concurrently. 
Similar to~\citep{schulman2017proximal,heess2017emergence}, we also use an adaptive-learning rate based on the KL divergence between the current and previous policies. 
Additionally, for stability we whiten the current observations by maintaining online statistics of mean and standard deviation from the history of past observations. 

Our policy and value functions share the same feed-forward network architectures.
As in the Baselines implementation of PPO, we use scaled exponential linear units (SELU ~\citep{klambauer2017self}) for the activation function. 
SELU implicitly encourages normalized activations, and in our hyperparameter search, policies with SELU learned faster and achieved higher rewards than those with ReLU and Tanh. 

For our multi-GPU benchmarks, we implemented two variants of Distributed PPO. 
Our first variant uses multiple GPUs to generate rollouts but trains the policy and value function on a single GPU.
This is often the case with CPU based implementations, where each CPU worker generates rollouts, and one master CPU trains. 
Our second variant is similar to Distributed PPO~\citep{heess2017emergence}. 
Both are synchronized algorithms, where at every iteration, gradients from all workers are applied to a central policy, and the worker policies are updated with the new policy parameters and this is what we use for all our experiments. 
We found that the first variant was not scalable to multiple nodes.
We also experimented with averaging parameters, but we found that it performed significantly worse than averaging gradients. 
We use Horovod~\footnote{\url{https://github.com/uber/horovod}} for distributed simulation and training across multiple GPUs and nodes where each GPU runs its own simulation and training instance in Tensorflow. 
The weight parameters are updated by averaging gradients across multiple GPU workers using efficient allreduce by NCCL~\footnote{\url{https://developer.nvidia.com/nccl}} and broadcasting from a master GPU. 
Importantly, since the \textit{env.step()} function is implemented directly on GPU for multiple agents, we are able to leverage the parallelism offered by GPUs to obtain observation-action pairs concurrently for hundreds to thousands of agents. 



\subsection{Hardware}
All experiments were done using NVIDIA Tesla V100 GPUs on NVIDIA's internal compute cluster.
Each single-GPU experiment uses 1 CPU core of a 20-Core Intel Xeon E5-2698 v4 processor running at 2.2 GHz.
For multi-GPU experiments, we scale the number of CPU cores used to match the number of GPUs used.

\subsection{Single-GPU Simulation and Training}
We first performed hyperparameter grid-search on the number of frames used per PPO optimization iteration and the network architectures for training 1024 parallel agents on the Ant and Humanoid tasks.
For both tasks, we found the best policy and value network architectures to have 3 hidden layers decreasing in size.
See Appendix~\ref{appendix:hyperparams} for the specific architectures and hyperparameters used.

We found the best frames per iteration within our searched values for Ant and Humanoid to be $32$ frames per iteration $\times 1024$ agents = $32768$ frames. 
This number is kept constant as we scale the number of parallel agents simulated up and down.
For example, for $512$ agents, we use $64$ frames per iteration, and for $2048$ we use $16$.
Keeping the frames per iteration constant helps us to show differences in learning speed as caused by improvements in simulation speed, and not by performance of the learning algorithm. 
We note that for high agent counts, the small number of frames used still enable learning, because our tasks use dense rewards.

\begin{figure*}[!htb]
\vspace{2mm}
\centerline{
\hfill { \includegraphics[width=0.33\linewidth]{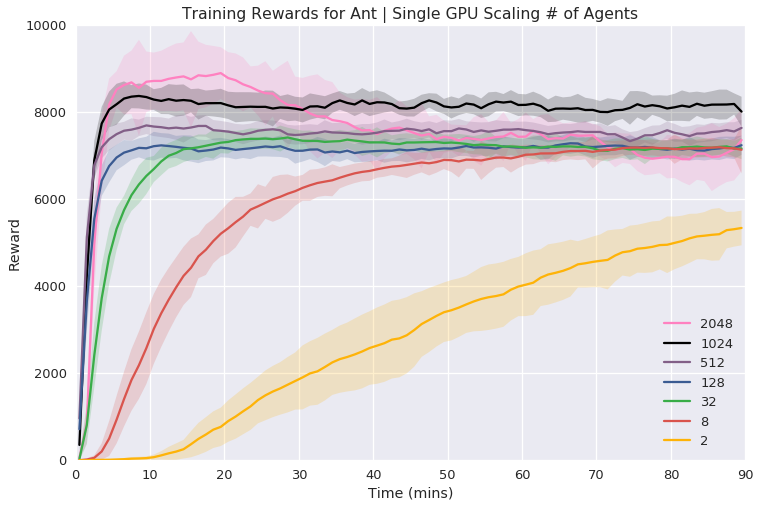} }
\hfill { \includegraphics[width=0.33\linewidth]{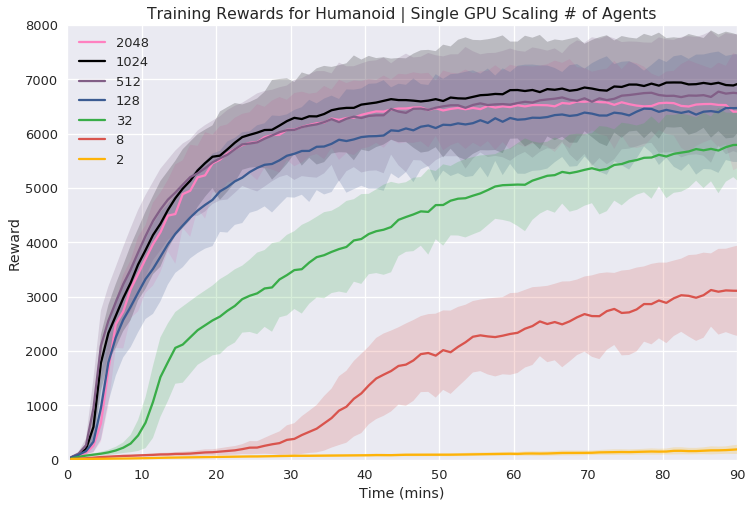} }
\hfill { \includegraphics[width=0.33\linewidth]{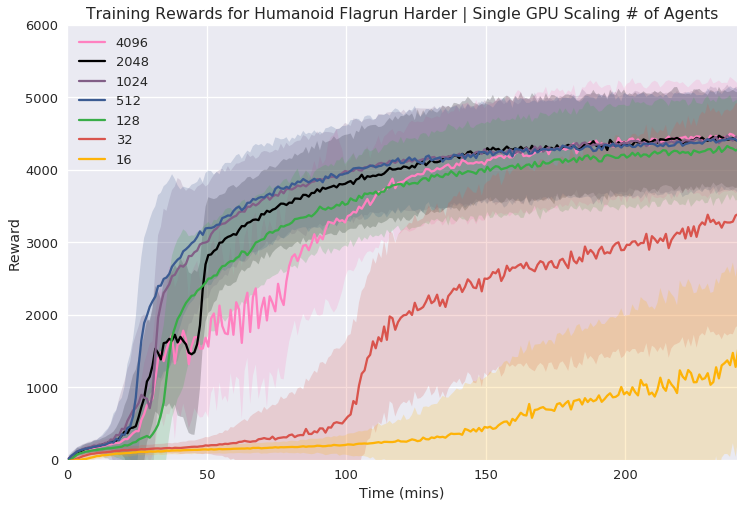} }
\hfill
}
\centerline{
\hfill { \includegraphics[width=0.325\linewidth]{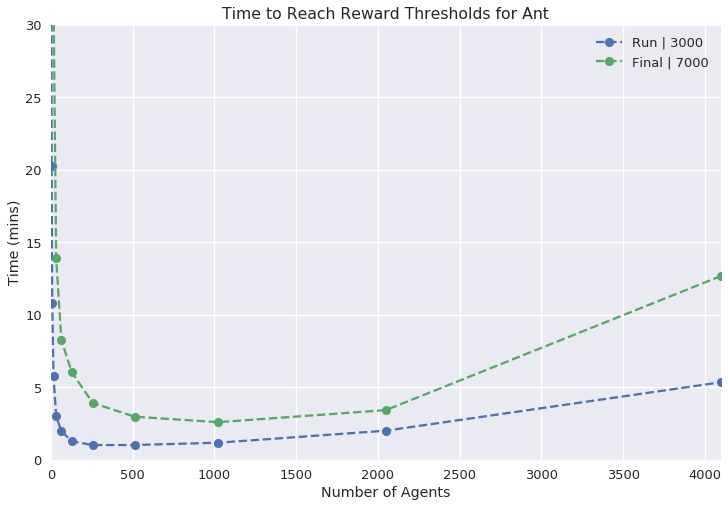} }
\hfill { \includegraphics[width=0.325\linewidth]{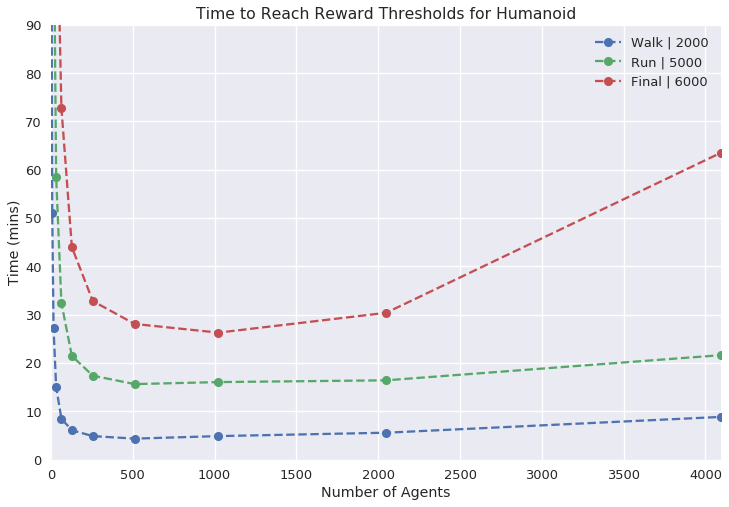} }
\hfill { \includegraphics[width=0.325\linewidth]{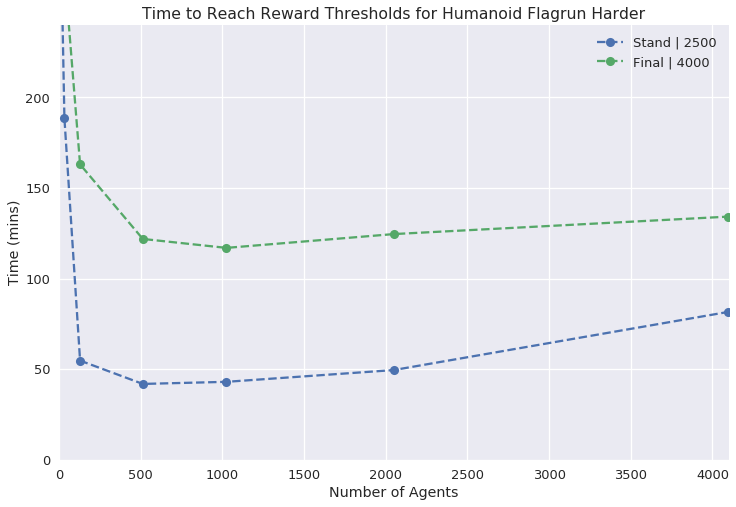} }
\hfill
}
\centerline{
\hfill \makebox[0.33\linewidth][c] {\footnotesize (a) Ant}
\hfill \makebox[0.33\linewidth][c] {\footnotesize (b) Humanoid}
\hfill \makebox[0.33\linewidth][c] {\footnotesize (c) Humanoid Flagrun Harder}
\hfill
}
\caption{\footnotesize Single GPU Experiments. We show the reward curves and wall time needed for training various tasks using increasing numbers of simulated agents to reach certain reward thresholds. The number of agents we evaluated vary in the powers of 2. The reward thresholds were chosen for significant behavior changes. For example - at around 2500 reward the Humanoid Flagrun Harder agents learn to stand up from sitting positions and begin walking. We keep the amount of experience used per PPO update iteration constant across evaluations by decreasing the frames used per agent as the number of agents increase. Using 512 agents we were able to train the Humanoid agent to run in about 16 minutes. All experiments were running against the same set of seeds for consistent comparison.}
\label{fig:single_gpu_scaling}
\end{figure*}

We report the time needed to reach certain reward thresholds for the Ant, Humanoid, and HFH tasks as we vary the number of agents in Figure~\ref{fig:single_gpu_scaling}.
Because we fix the amount of experience used per PPO update constant, we are able to observe a trade-off between increasing the number of agents but collecting less frames per agent and decreasing the number of agents but collecting more frames per agent.
The point of diminishing return varies across task and reward thresholds. 

We also note the short time needed to learn these tasks using GPU-accelerated simulations.
We list training times and resources used for the humanoid task in prior works in Table~\ref{tab:humanoid_times}, all of which used CPU-based physics simulation.
With one GPU and CPU core, the Humanoid agents were able to run in less than 20 minutes while using 10$\times$ to a 1000$\times$ less CPU cores than previous works~\citep{mania2018simple, salimans2017evolution}.

\begin{table}[h!]
  \begin{center}
    \begin{tabular}{c|c|c|c} 
      \textbf{Algorithm} & \textbf{CPU Cores} & \textbf{GPUs} & \textbf{Time (mins)}\\
      \hline
      Evolution Strategies~\cite{salimans2017evolution}                      & 1440 & - & 10 \\
      Augmented Random Search~\cite{mania2018simple}                         & 48   & - & 21 \\
      Distributed Prioritized Experience Replay~\cite{horgan2018distributed} & 32   & 1 & 240 \\
      Proximal Policy Optimization w/ GPU Simulation (Ours)                  & 1    & 1 & 16  \\ 
      \hline
    \end{tabular}
    \caption{ \footnotesize Resources and Times for Training a Humanoid to Run. Prior works all used CPU-based physics simulations. In this table, we do not include the original PPO paper~\citep{schulman2017proximal} - it used 128 CPU cores for the humanoid task but did not report training time. We also did not include the Distributed PPO paper~\citep{heess2017emergence} - their humanoid training took more than 40 hours to converge, but they did not report the number of CPU cores used.}
    \label{tab:humanoid_times}
  \end{center}
  \vspace{-4mm}
\end{table}

\subsection{Multi-GPU Distributed Simulation and Training}
\begin{figure*}[!htb]
\vspace{2mm}
\centerline{
\hfill { \includegraphics[width=0.33\linewidth]{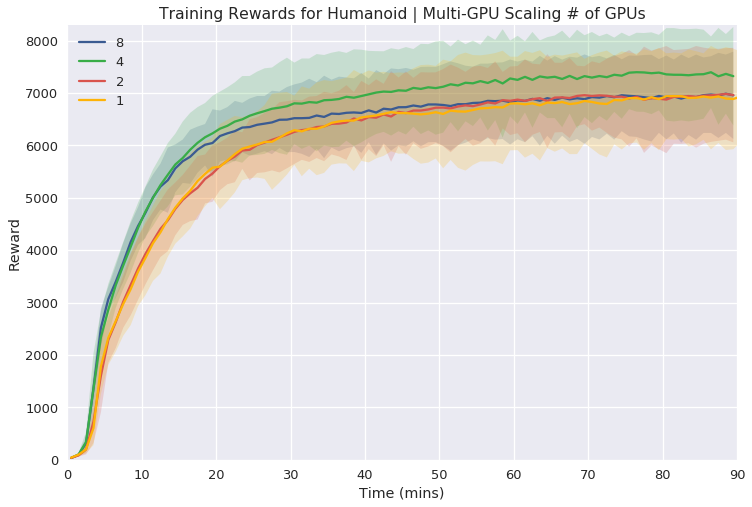} }
\hfill { \includegraphics[width=0.33\linewidth]{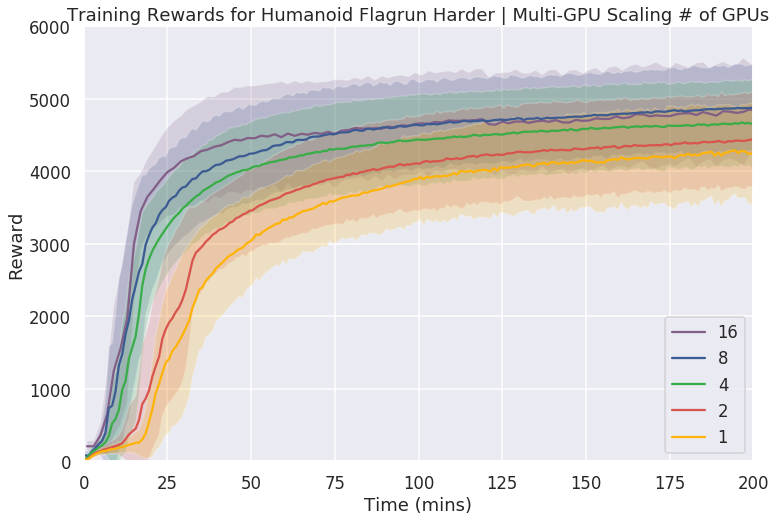} }
\hfill { \includegraphics[width=0.33\linewidth]{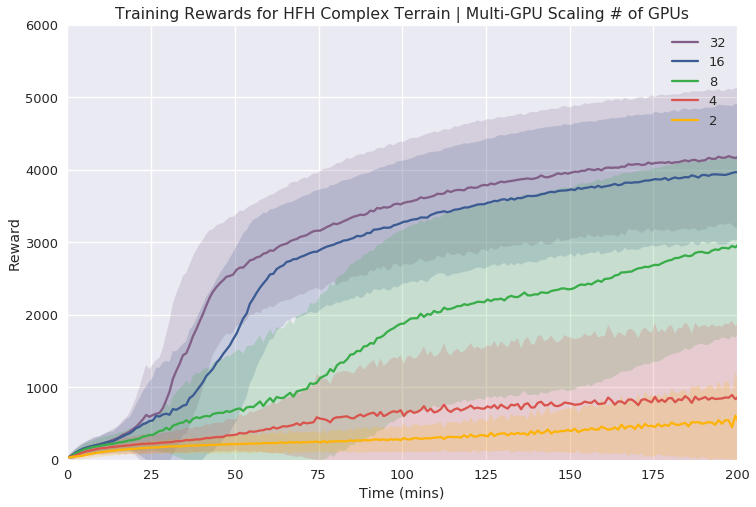} }
\hfill
}\centerline{
\hfill \makebox[0.33\linewidth][c] {\footnotesize (a) Humanoid}
\hfill \makebox[0.33\linewidth][c] {\footnotesize (b) Humanoid Flagrun Harder (HFH)}
\hfill \makebox[0.33\linewidth][c] {\footnotesize (c) HFH Complex Terrain}
\hfill
}
\caption{\footnotesize Multi-GPU Simulation and Training. We show how our GPU-accelerated RL simulation can be scaled to simulation and training with multiple GPUs. The overlap of standard deviations indicates that there are little scaling effects for the Humanoid task, and the benefit of multi-GPU simulation and training is only apparent for more complex tasks and with a greater difference in the number of GPUs used. All experiments were running against the same set of seeds for consistent comparison.}
\label{fig:multi_gpu_envs}
\vspace{-5mm}
\end{figure*}

We extend our method to distribute GPU simulations across multiple GPU workers to see how learning speed can be improved on the Humanoid, HFH, and HFH on Complex Terrain tasks. 
For these experiments, we run a simulation and training instance on each GPU, and we use Horovod for distributed gradients averaging.
We also normalize the advantage estimates across all GPUs and distribute them back to each GPU worker at every iteration, ensuring that advantages across all GPUs share a consistent global mean and standard deviation.
The number of agents simulated per GPU for Humanoid and HFH is 1024.
We use a smaller number of 512 agents per GPU for HFH on Complex Terrain to keep memory usage and simulation speed reasonable, as the addition of the height map significantly increases the dimensionality of the observations.
Results are reported in Figure~\ref{fig:multi_gpu_envs}.
We observe only limited scaling effects for the Humanoid task, which hit diminishing returns after 4 GPUs.
In the more complex tasks however, we observed noticeable speed-ups.
For the HFH task, the 1 GPU run reached 4000 rewards in about 2 hours, while the 8 GPU run reached it in about 30 minutes, and 16 GPUs in about 15.
For the HFH on Complex Terrain task, we observe more apparent scaling benefits with multiple GPU simulation and training.
On average, the 16 and 32 GPU runs learn the task faster than the 2, 4, and 8 GPU runs, while the large overlap in standard deviations for 8 GPUs with 16 and 32 shows the diminishing returns of using more agents in learning. 

\section{Conclusion and Future Work}
\label{sec:future-works-conclusion}

In this work, we used an in-house GPU-accelerated physics simulator to concurrently simulate hundreds to thousands of robots for Deep RL of continuous-control, locomotion tasks.
In contrast to prior works that trained locomotion tasks on CPU clusters, with some using hundreds to thousands of CPU cores, we are able to train a humanoid to run in less than 20 minutes on a single machine with 1 GPU and CPU core, making GPU-accelerated RL simulation a viable alternative to CPU-based ones.
Our RL simulation framework can also be scaled to multi-GPU and multi-node settings, and we observed that multi-GPU simulation and training shows greater learning speed improvements for more complex locomotion tasks.
Given the recent successes of sim2real transfer learning, from grasping in clutter~\cite{mahler2017learning}, quadruped locomotion~\cite{tan2018sim}, to dexterous manipulation~\cite{1808.00177}, all of which used Bullet to generate simulation data to train policies that worked in the real world, we believe our simulator can provide valuable speed-ups for similar applications in the future. 

In future work, we plan to experiment with more complex humanoid environments by allowing the humanoid to actively control the orientation of the rays used to generate the height map.
This may enable the humanoids to navigate dynamic obstacles and obstacles in mid-air.
We also plan to use our simulator for manipulation tasks with robots such as the Fetch, Baxter, and YuMi.
In this work, we've considered locomotion tasks with full state information.
For many tasks in manipulation and navigation however, training from vision data is preferred. 
For such tasks, we note the potential of zero-copy training - directly feeding simulation data generated by a GPU-based simulator and the task's states and rewards into a deep learning framework without the data leaving the GPU.
Zero-copy training eliminates the need to communicate data from the GPU to the CPU, and can further improve training speed.


\clearpage
\acknowledgments{We thank Phil Rogers, Vikrama Ditya, Christopher Lamb, Nathan Luehr, David Addison, Hari Sundararajan, Sivakumar Arayandi Thottakara, Julie Bernauer and many others who manage the NVIDIA GPU infrastructure for all the kind help they provided in carrying out the experiments on the GPU clusters.}


\bibliography{refs}  


\newpage
\appendix

\section{Rewards vs Frames}
\label{appendix:frames}

\begin{figure*}[!htb]
\vspace{2mm}
\centerline{
\hfill { \includegraphics[width=0.33\linewidth]{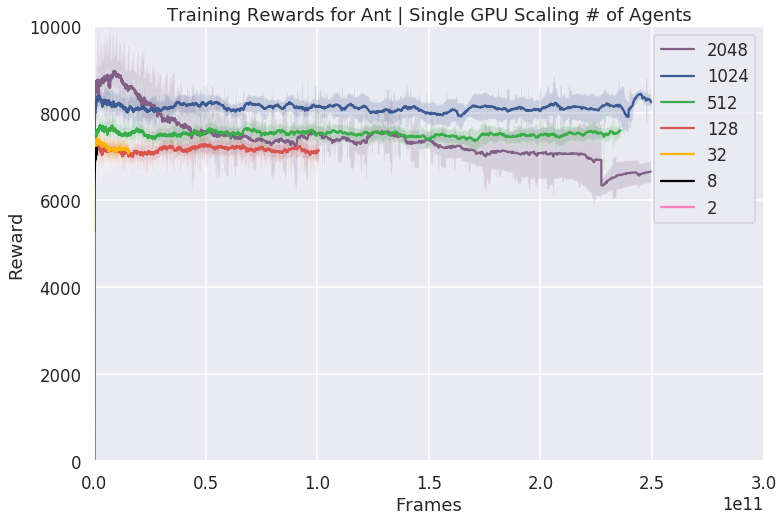} }
\hfill { \includegraphics[width=0.33\linewidth]{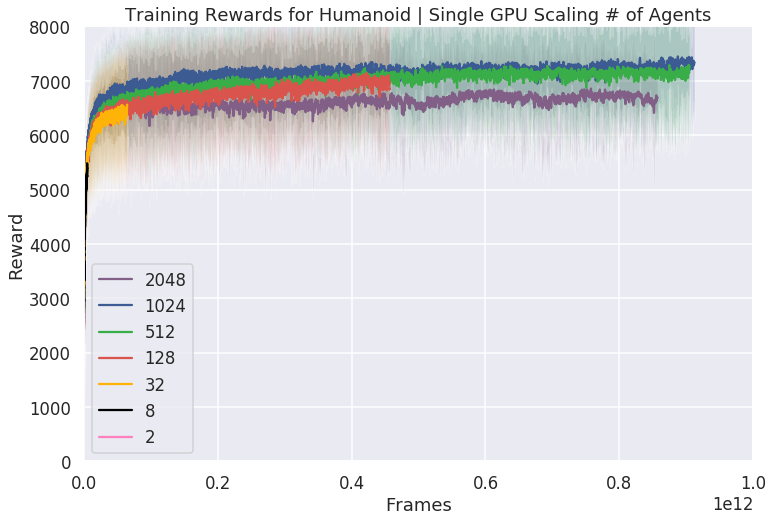}}
\hfill { \includegraphics[width=0.33\linewidth]{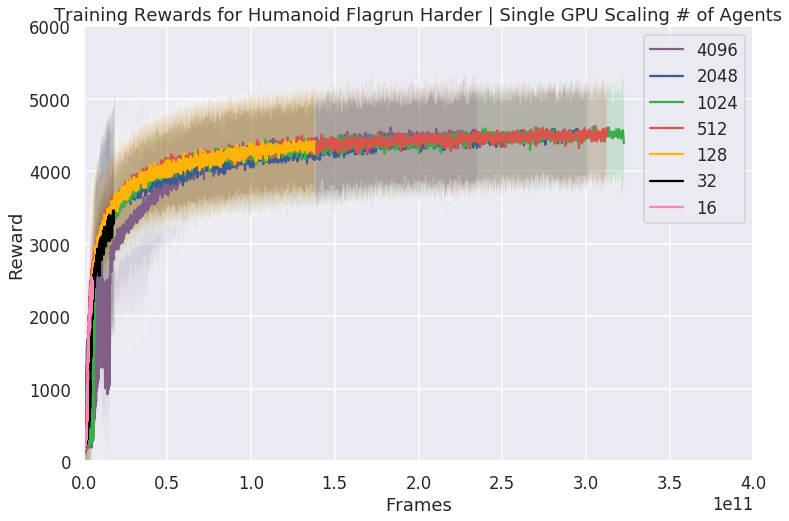} }
\hfill
}
\centerline{
\hfill { \includegraphics[width=0.325\linewidth]{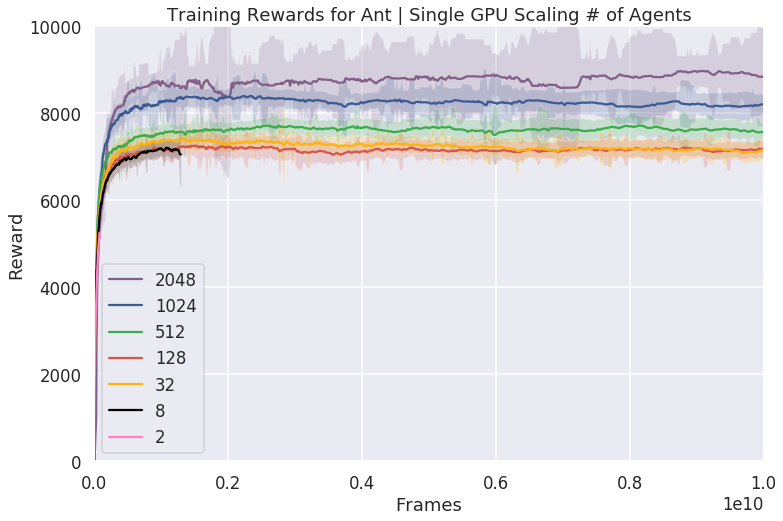} }
\hfill { \includegraphics[width=0.325\linewidth]{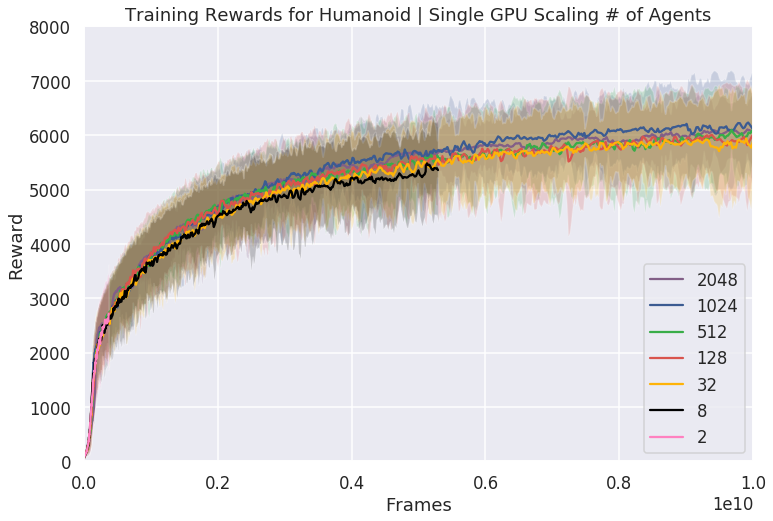} }
\hfill { \includegraphics[width=0.325\linewidth]{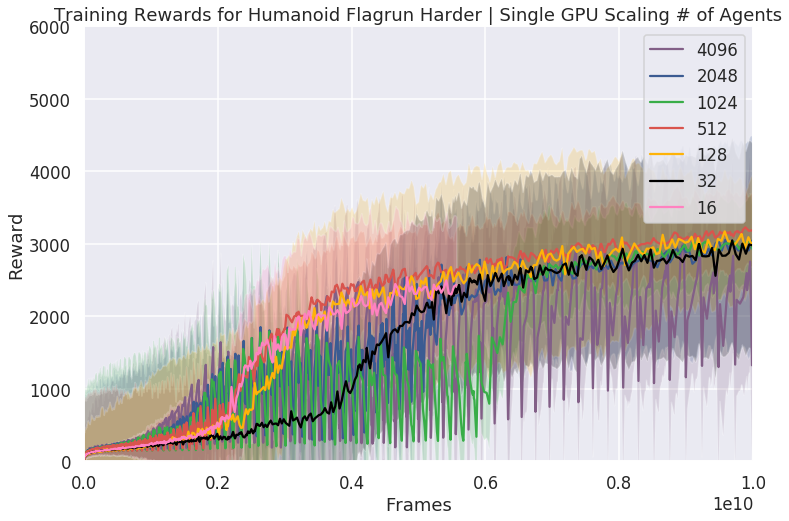} }
\hfill
}
\centerline{
\hfill \makebox[0.33\linewidth][c] {\footnotesize (a) Ant}
\hfill \makebox[0.33\linewidth][c] {\footnotesize (b) Humanoid}
\hfill \makebox[0.33\linewidth][c] {\footnotesize (c) Humanoid Flagrun Harder}
\hfill
}
\caption{\footnotesize Reward vs Frames for Single GPU Experiments.}
\label{fig:single_gpu_scaling_frames}
\end{figure*}

\begin{figure*}[!htb]
\vspace{2mm}
\centerline{
\hfill { \includegraphics[width=0.33\linewidth]{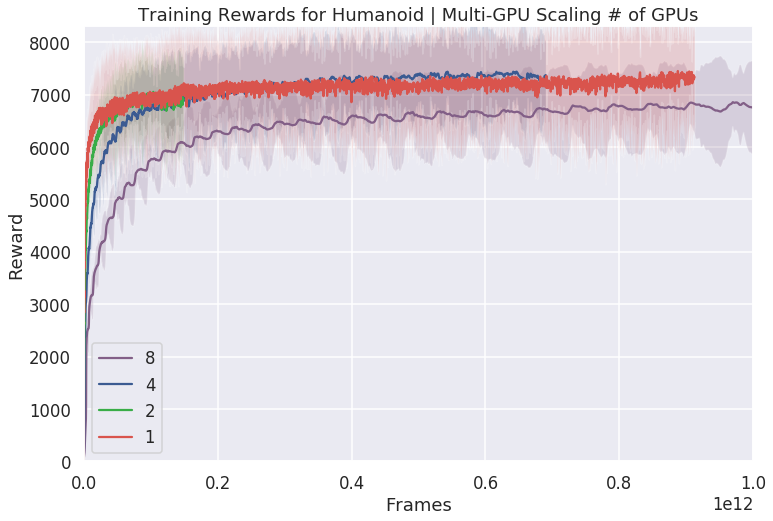} }
\hfill { \includegraphics[width=0.33\linewidth]{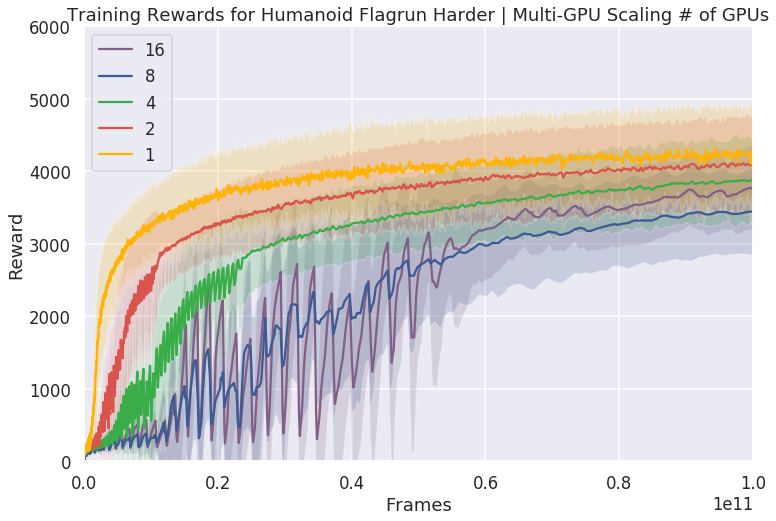}}
\hfill { \includegraphics[width=0.33\linewidth]{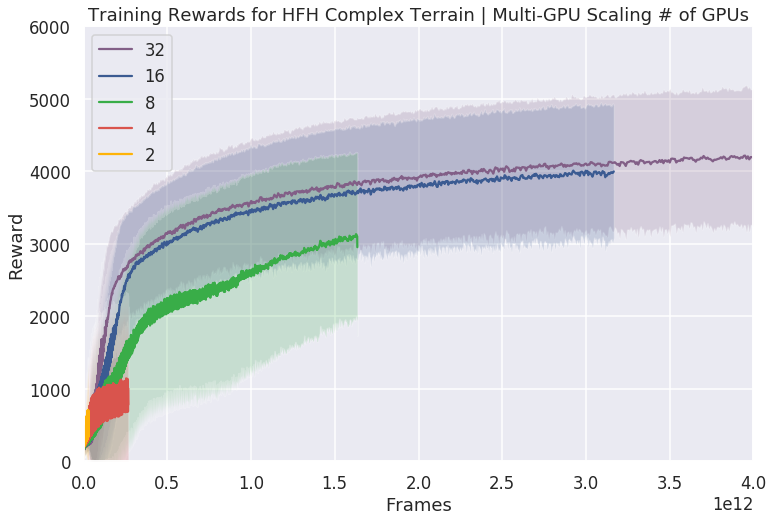} }
\hfill
}
\centerline{
\hfill { \includegraphics[width=0.325\linewidth]{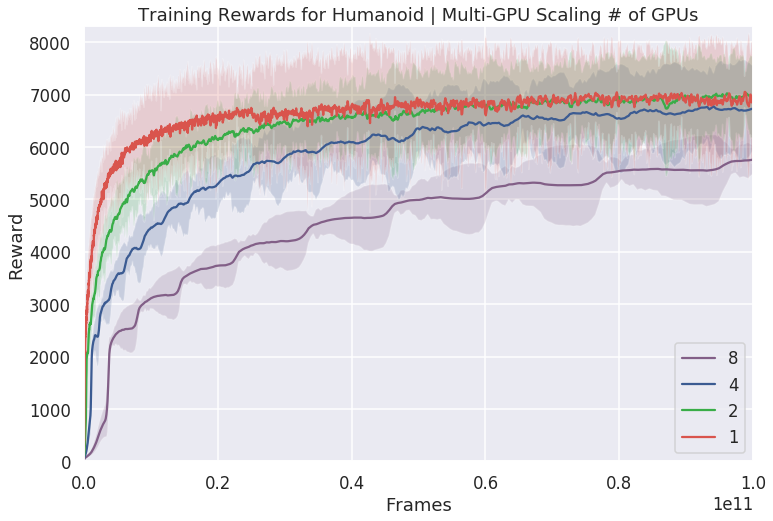} }
\hfill { \includegraphics[width=0.325\linewidth]{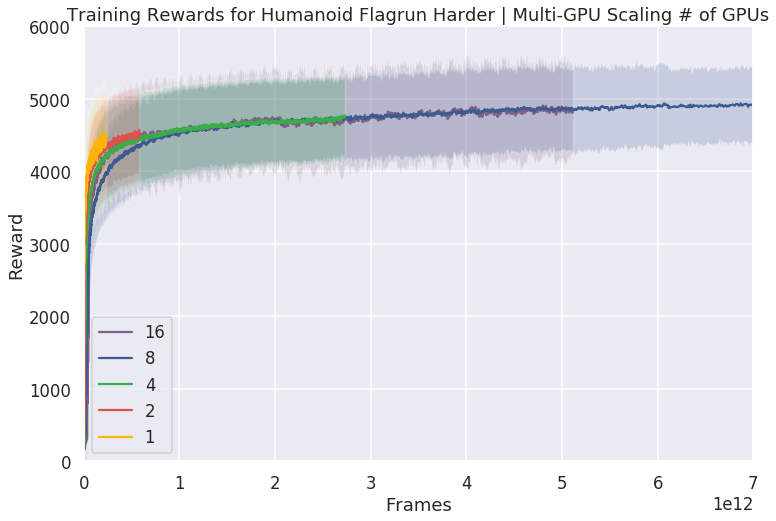} }
\hfill { \includegraphics[width=0.325\linewidth]{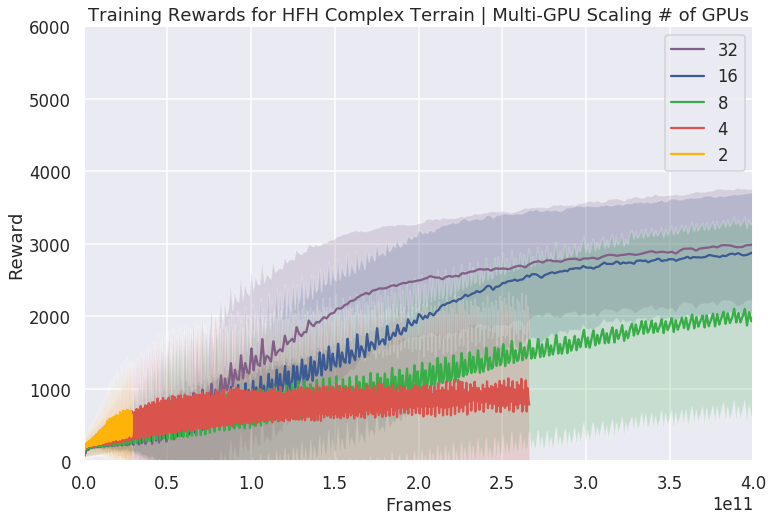} }
\hfill
}
\centerline{
\hfill \makebox[0.33\linewidth][c] {\footnotesize (a) Humanoid}
\hfill \makebox[0.33\linewidth][c] {\footnotesize (b) Humanoid Flagrun Harder (HFH)}
\hfill \makebox[0.33\linewidth][c] {\footnotesize (c) HFH Complex Terrain}
\hfill
}
\caption{\footnotesize Reward vs Frames for Multi-GPU Experiments.}
\label{fig:multi_gpu_scaling_frames}
\end{figure*}

We plot the reward vs frames curves for single-GPU experiments in Figure~\ref{fig:single_gpu_scaling_frames} and multi-GPU experiments in Figure~\ref{fig:multi_gpu_scaling_frames}.
A zoomed-in version of each plot is shown on the second row.
The difference in the number of total frames for different number of agents is due to the fact that we stop training based on a fixed amount of time.

\newpage
\section{Comparison of Observations}
\label{appendix:obs_comp}
Table~\ref{tab:state} compares the different observations for the Humanoid running task used in MuJoCo, Roboschool, Control Suite, and our environments. 

\begin{table}[!htb]
\def\arraystretch{1.1}%
\begin{tabular}{l|lllll}
                              & MuJoCo~\footnote{\url{https://github.com/openai/gym/wiki/Humanoid-V1}}             & Roboschool~\footnote{\url{https://github.com/openai/roboschool/blob/master/roboschool/gym_forward_walker.py}} & Control Suite~\footnote{\url{https://github.com/deepmind/dm_control/blob/master/dm_control/suite/humanoid.py}} & Ours        &  \\ \cline{1-5}
Root Body Height              & $1$                  & $1$                & $1$           & $1$          &  \\
Root Body Rotation            & $4$                  & $2$                & $3$           & $2$          &  \\
Root Body Velocity            & -                    & $3$                & $3$           & $3$          &  \\
Root Body Angular Velocity    & -                    & -                  & -             & $3$          &  \\
Root Body Heading Direction   & -                    & $2$                & -             & $2$          &  \\
Joint Angles                  & $17$                 & $17$               & $21$          & $21$         &  \\
Joint Angle Velocities        & $23$                 & $17$               & -             & $21$         &  \\
Positions of Hands and Feet    & -                    & -                  & $4\times3$    & -            &  \\
Velocities of Bodies          & $14 \times 3$        & -                  & $27$          & -            &  \\
Angular Velocities of Bodies  & $14 \times 3$        & -                  & -             & -            &  \\
Inertia Tensor and Mass       & $14 \times 10$       & -                  & -             & -            &  \\
Actuator Forces               & $23$                 & -                  & -             & $21$         &  \\
External Forces on Bodies     & $14 \times 6$        & -                  & -             & -            &  \\
Whether Feet in Contact w/ Ground & -                & $2$                & -             & $2$          &  \\\cline{1-5}
\textbf{Total}                & $376$                & $44$               & $67$          & $76$         &  \\
\end{tabular}
\caption{\footnotesize Observation and dimensionality comparison for Humanoid running task. The Root Body Heading Direction for Roboschool and our agents are represented by the sine and cosine values of the angular difference between our agent's heading and the direction to the target (which for the humanoid running task is just forwards from the starting location). The Root Body Rotation is represented as quaternions for MuJoCo, roll and pitch angles for Roboschool and ours, and the z-projection of the rotation matrix for Control Suite.}
\label{tab:state}
\end{table}

\section{Rewards}
\label{appendix:rewards}
The reward function we used for all four tasks are as follows:

\begin{equation}
R = R_\text{alive} + S + 0.5R_\text{heading} + 0.05R_\text{standing} - 4\frac{1}{\tau_\text{max}}||\tau||_1 - 0.5||u||_2^2 - 0.2N_\text{joints} - N_\text{feet}
\end{equation}

\begin{equation} 
    R_\text{heading} = \begin{cases} 
      1 & \cos(\theta_\text{target}) > 0.8 \\
      \cos(\theta_\text{target})/0.8   & \cos(\theta_\text{target}) <= 0.8
   \end{cases}
\end{equation}
\begin{equation}
R_\text{standing} = \mathds{1}\{\cos(\theta_\text{vertical}) > 0.93\}
\end{equation}

$\theta_\text{target}$ is the angle from the robot's current heading to the angle toward the target location.
$\theta_\text{vertical}$ is the angle of the robot's torso from the vertical-axis (i.e. if the humanoid is standing up right, this angle would be 0). 
$R_\text{alive}$ is the alive bonus, and it is $0.5$ for Ant and $2$ for humanoids.
$S$ is the speed toward the current target.
$\tau$ is the vector of motor torques applied at each joint, with $\tau_\text{max}$ being the maximum that can be applied.
$u$ is the current action. 
$N_\text{joints}$ is the number of joints at joint limits, and $N_\text{feet}$ is the number of feet that is in collision with ground. 

\newpage
\section{Comparison of Reward Functions}
\label{appendix:reward_comp}
Table~\ref{tab:rew_hum} compares the coefficients for the summands in the reward function for the Humanoid running task used in MuJoCo, Roboschool, and our environments. 

We don't list the reward function coefficients for the humanoid walking task in Deepmind Control Suite, as it is very different in structure - it is the product of the running speed with two coefficients that depend on the magnitude of the controls and how upright the humanoid is. 
See here~\footnote{\url{https://github.com/deepmind/dm_control/blob/master/dm_control/suite/humanoid.py}} for details.

\begin{table}[!htb]
\centering
\def\arraystretch{1.1}%
\begin{tabular}{l|l|l|ll}
                        & MuJoCo~\footnote{\url{https://github.com/openai/gym/blob/master/gym/envs/mujoco/humanoid.py}} & Roboschool~\footnote{\url{https://github.com/openai/roboschool/blob/master/roboschool/gym_forward_walker.py}} & Ours                        &  \\ \cline{1-4}
Alive Bonus             & $5$                & $2$                 & $2$                         &  \\
Running Speed Bonus     & $0.25$             & $1$                 & $1$                         &  \\
Heading Bonus           & -                  & -                   & $0.5$                       &  \\      
Standing Bonus          & -                  & -                   & $0.05$                      &  \\      
Control Cost            & $-0.1$             & $-\frac{0.1}{N_d}$  & $-0.5$                      &  \\ 
Electricity (Torque) Cost       & -          & $-4.25$                & $-\frac{4}{\tau_\text{max}}$&  \\
Joints at Limits Cost   & -                  & $-0.2$              & $-0.2$                      &  \\
Feet Contact Cost       & -                  & $-1$                & $-1$                        &  \\ 
External Forces Cost    & $-5\times10^{-6}$ & -                   & -                           &  \\ 
\hline
\end{tabular}
\caption{\footnotesize 
Reward function comparison for Humanoid Running task. $N_d$ is the dimension of the controls. 
The External Forces Cost for MuJoCo is the multiplier of the sum of squares of external forces on all bodies. 
}
\label{tab:rew_hum}
\end{table}

\section{Hyperparameters}
\label{appendix:hyperparams}

In table~\ref{tab:hyperparams} we give the hyperparameters used during training for Ant, Humanoid, and Humanoid Flagrun Harder tasks. 
The timesteps per batch is given relative to a specific amount of parallel agents simulated - $1024$, and this is scaled across different experiments. 
For example, if the timesteps per batch is $32$, then we used $64$ with the experiment that has $512$ agents, and $16$ with the experiment that has $2048$ agents.
The desired KL specifies the target KL value used for adapting the Adam step size at each iteration.

The value and policy networks in our PPO algorithm have the same feed-forward architectures, and they are specified in a list format, where the $n$th value gives the size of the $n$th layer.

\begin{table}[!htb]
\centering
\def\arraystretch{1.1}%
\begin{tabular}{l|l|l|l|l}
Hyperparameter           & Ant  & Humanoid & HFH & HFH Terrain\\ \hline
Timesteps per Batch      & 32   & 32 & 64 & 64       \\
Num. Epochs              & 20   & 20 & 10 & 20        \\
Minibatch Size per Agent & 16   & 32 & 32 & 8        \\
Desired KL               & 0.01 & 0.02  & 0.01  & 0.01     \\
Neural Net Architecture  & [128, 64, 32] & [256, 128, 64] & [512, 256, 128] & See Caption \\
\end{tabular}
\caption{\footnotesize Hyperparameters used in different tasks. For the HFH Terrain neural network, we pass the $15 \times 11$ height map into two fully-connected (FC) layers of sizes $[256, 128]$, the other observations through one layer of size $512$, then finally pass their concatenated outputs through two more FC layers of sizes $[256, 128]$ before outputting the controls.}
\label{tab:hyperparams}
\end{table}

\newpage
\section{MuJoCo Simulation Times}
\begin{figure}[!htb]
    \centering
    \includegraphics[width=0.46\linewidth]{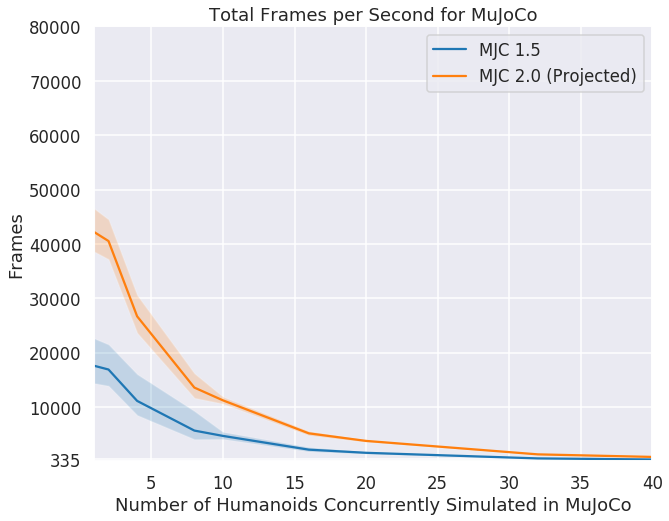}
    \includegraphics[width=0.44\linewidth]{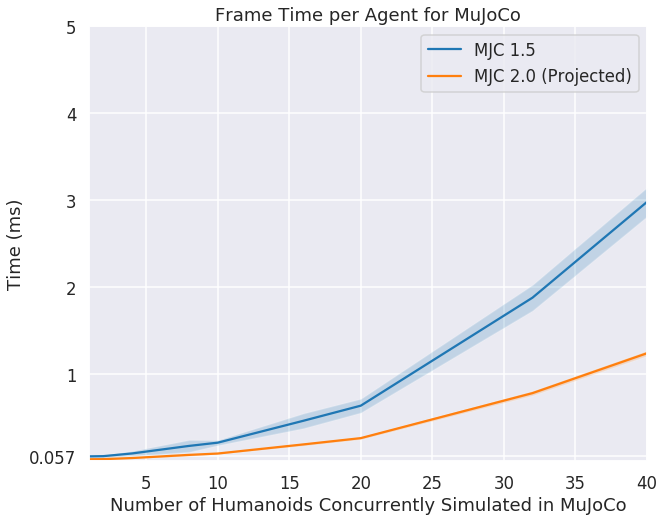}
    \caption{\footnotesize MuJoCo Simulation Speed. }
    \label{fig:mjc_sim_times}
\vspace{-4mm}
\end{figure}

In Figure~\ref{fig:mjc_sim_times} we report plots similar to those in Figure~\ref{fig:sim_times} where we evaluate the total frames per second and frame time per agent on MuJoCo 1.5 as the number of concurrent humanoids simulated in the scene increases.
We measured MuJoCo's single-core CPU simulation time in a similar setup as we did with our GPU simulation time - at every time step random actions are given to the $28$-DoF humanoids lying on the floor.
The CPU used is an Intel Core i9-7960X running at 2.80GHz.
At the time of writing MuJoCo 2.0 has just been released, but it is not yet supported by mujoco\_py. 
We plot the projected curves for MuJoCo 2.0 by reducing the simulation time of MuJoCo 1.5 by $40\%$, as reported here~\footnote{\url{https://www.roboti.us/index.html\#mujoco200}}.

We note that while MuJoCo performs well on one core with one humanoid, the simulation time increases as the number of humanoids in the scene increases.
This is in contrast to GPU simulations, where having more concurrent simulations with more contacts, which enable large-scale inter-agent interactions, reduces the simulation time per agent. 

\end{document}